# OpenMedLM: Prompt engineering can out-perform fine-tuning in medical question-answering with open-source large language models

Jenish Maharjan[a]*, MS, Anurag Garikipati[a]*, MS, Navan Preet Singh[a], MS, Leo Cyrus[a], PhD, Mayank Sharma[a], MS, Madalina Ciobanu[a], PhD, Gina Barnes[a], MPH, Rahul Thapa[a], BS, Qingqing Mao[a], PhD, Ritankar Das[a], MSc

*Joint first authors

Author Affiliations

[a]Montera, Inc dba Forta

548 Market St, PMB 89605

San Francisco, CA 94104-5401

United States

**Abstract**

**Background**: LLMs have become increasingly capable at accomplishing a range of specialized-tasks and can be utilized to expand equitable access to medical knowledge. Most medical LLMs have involved extensive fine-tuning, leveraging specialized medical data and significant, thus costly, amounts of computational power. Many of the top performing LLMs are proprietary and their access is limited to very few research groups. However, open-source (OS) models represent a key area of growth for medical LLMs due to significant improvements in performance and an inherent ability to provide the transparency and compliance required in healthcare. Here, we present OpenMedLM, a prompting platform which delivers state-of-the-art (SOTA) performance for OS LLMs on medical benchmarks.

**Methods**: We evaluated a range of OS foundation LLMs (7B-70B) on four medical benchmarks (MedQA, MedMCQA, PubMedQA, MMLU medical-subset). Yi 34B performed the best at baseline, and thus was utilized for developing OpenMedLM. We employed a series of prompting strategies, including zero-shot, few-shot, chain-of-thought (random selection and kNN selection), and ensemble/self-consistency voting.

**Results**: Through a series of robust prompt engineering techniques, we found that OpenMedLM delivers OS SOTA results on three common medical LLM benchmarks, surpassing the previous best performing OS models that leveraged computationally costly extensive fine-tuning. OpenMedLM displays the first results to date that demonstrate the ability of OS foundation models to significantly optimize performance, absent specialized fine-tuning. The model delivers a 72.6% accuracy on the MedQA benchmark, outperforming the previous SOTA by 2.4%, and achieves 81.7% accuracy on the MMLU medical-subset, establishing itself as the first OS LLM to surpass 80% accuracy on this benchmark.

**Conclusion**: Our results highlight medical-specific emergent properties in OS LLMs which have not yet been documented to date elsewhere, and showcase the benefits of further leveraging prompt engineering to improve the performance of accessible LLMs for medical applications.

# 1. Introduction

The rise in the popularity of large language models (LLMs) since the release of ChatGPT ([Anon], 2015-2024) has been met with significant amounts of excitement for their potential applications. There has been particular enthusiasm for exploring the integration of generalist LLMs and other generative artificial intelligence (AI) tools in highly specialized and complex tasks, such as those related to healthcare/medicine (Thirunavukarasu et al., 2023), software programming ([Anon], 2023b), and music creation (Agostinelli et al., 2023), resulting in an array of specialized models. LLMs are continuously upgraded with new use-cases being ongoingly developed, and the accuracy of these models is showcased through their performance on various benchmark tasks, including multitask multiple choice question and answering (Massive Multitask Language Understanding [MMLU]) (Hendrycks et al., 2020), natural language inference (HellaSwag) (Zellers et al., 2019), and problem solving (BIG-Bench) (Srivastava et al., 2022). Many of the best performing LLMs available today are proprietary models with hundreds of billions of parameters that were trained on hundreds of billions or trillions of tokens (Hoffmann et al., 2022). For example, proprietary generalist models, such as GPT-3 (Brown et al., 2020), GPT-4 ([Anon], March 27, 2023), PaLM (Chowdhery et al., 2022), and Gemini (Anil et al., 2023), report training on massive compute infrastructure to utilize the huge number of tokens available for their training and the massive number of parameters encompassing the model, thereby ensuring performance optimization. As these models grow larger, they increasingly display very high accuracy on standard LLM benchmarks, with GPT-4 achieving state-of-the-art (SOTA) performance on measures such as HellaSwag (Zellers et al., 2019) and Gemini realizing SOTA performance on multiple benchmarks including MMLU and BIG-Bench (Anil et al., 2023).

Open-source (OS) models have also shown promise in achieving comparable performance to proprietary models, while providing the advantages of increased flexibility, compliance benefits, and task specification capabilities. In particular, OS models can enable developers to access model weights directly, affording the ability to host models on a user's own infrastructure, perform highly granular fine-tuning, and accomplish tasks which may not be possible through a secondary platform or application programming interface (API), such as querying local databases to complete specific tasks through retrieval-augmented generation (Lewis et al., 2020). Generalist OS LLMs such as Meta's Llama ([Anon], 2023d) and Llama-2 ([Anon], 2023c) suite of models, Mistral AI's Mistral (Jiang et al., 2023) and Mixtral (Jiang et al., 2024), and 01.AI's Yi ([Anon], 2023a) suite of models have achieved very strong performance metrics across various benchmarks despite employing only a fraction of the parameters used by the larger, proprietary models. These generalist OS foundation models have enabled researchers, companies, and AI enthusiasts to train robust fine-tuned models with improved performance, for example task-specific models specialized for healthcare use (Toma et al., 2023; Wu et al., 2023).

LLMs have become increasingly more capable at a growing range of language tasks, and they inherently constitute a powerful tool that can expand equitable access to medical knowledge (Gottlieb & Silvis, 2023). This has promoted interest in the integration within various medical tasks of LLM-based AI tools, such as diagnostics aids, automation tools for administrative tasks, and note-taking tools designed for healthcare providers, as well as tools developed for patient use to help with understanding care plans and receiving personalized recommendations (Gottlieb & Silvis, 2023; Moor et al., 2023; Ramprasad et al., 2023; Thirunavukarasu et al., 2023). However, prior to deploying LLMs into real-world medical settings, models intended for healthcare implementation have to be proven to be accurate, unbiased, and safe for use with patients. As a first step to successfully developing healthcare LLMs, models have to be evaluated on medical benchmarks, which include assessing performance on multiple choice questions from a wide range of sources to determine accuracy on tasks with differing complexities and related to wide-ranging medical focus areas (Hendrycks et al., 2020; Jin et al., 2021; Jin et al., 2019; Pal et al., 2022). The interest in LLMs for medical tasks has led to an ever-increasing array of performance benchmarks for assessing healthcare suitability. Of particular interest to LLM-based healthcare tools are medical questions and answers (Q&A), which led to the development of specialized benchmarks for assessing model performance, accounting for questions from various medical licensing exams (Jin et al., 2021; Pal et al., 2022), questions on a corpora of biomedical research papers (Jin et al., 2019), and questions based on high school and college level life sciences topics (Hendrycks et al., 2020). Medical benchmarks that are typically used for performance evaluation of medically-specialized LLMs include MedQA (US Medical Licensing Exam [USMLE] questions), MedMCQA (Indian post-graduate medical exam), PubMedQA (Questions based on PubMed-indexed abstracts), and the medical-subset of the MMLU (nine medically and clinically relevant subsets of life sciences topics) (Hendrycks et al., 2020; Jin et al., 2021; Jin et al., 2019; Pal et al., 2022).

Optimizing performance of healthcare LLMs on medical benchmarks can entail a variety of approaches, with the majority involving some degree of fine-tuning, which may be accompanied by expensive computational costs that are out of reach for most and may require task-specific data that is not easily accessible, thus detracting from the goal of expanding equitable access to medical knowledge. Med-PaLM and Med-PaLM 2, which are proprietary specialized models released by Google, leverage instruction fine-tuning and instruction prompt tuning, with a focus on medical tasks, in order to improve upon the base models PaLM and PaLM 2, respectively (Chowdhery et al., 2022; Singhal, Tu, et al., 2023). While proprietary models have achieved strong performance, the ability to perform further investigation into the model's performance beyond published results is restrictive and limited to research groups with authorization to access the weights of the model. Adapting the tuning techniques used in proprietary model optimization to OS models can provide the broader research community with the ability to test and further improve upon results achieved by other groups, as well as with a deeper understanding of the mechanistic details behind these approaches. Furthermore, OS models offer the additional benefits of transparency and compliance, which is critical in a field with highly sensitive data, such as healthcare. OS specialized models, such as Clinical Camel and Meditron, have also leveraged fine-tuning techniques to boost performance over the corresponding OS foundation models (e.g., Llama-2 models) (Chen et al., 2023; Toma et al., 2023).

While fine-tuning has produced strong results on evaluating LLMs on medical benchmarks, recent studies have examined employing robust prompt engineering (a less costly option for model improvement) to optimize performance of generalist foundation models to similar or better levels when tested on medical benchmarks and compared with the performance of specialized fine-tuned models (Meskó, 2023; Nori et al., 2023). For example, Microsoft developed Medprompt, a robust prompting technique, for use with the generalist foundation GPT-4 model and achieved SOTA results on some of the most commonly used medical Q&A benchmarks (Nori et al., 2023). This highlights the potential of generalist LLMs to compete with specialized fine-tuned models for completing medical tasks while reducing the need for the expensive computational costs and other challenges (e.g., catastrophic forgetting, task-specific data acquisition) associated with fine-tuning specialized models. For example, performance optimization methods such as full-parameter fine-tuning suffer from catastrophic forgetting, which is a phenomenon where models lose some of their previously learned abilities as they learn new domain-specific information (Chen et al., 2020; Kirkpatrick et al., 2016; Korbak et al., 2022). This can potentially limit the ability to perform more complex tasks beyond the benchmark tasks evaluated during performance optimization.

Although specialized fine-tuned models leverage basic prompt engineering, there is no study to date showing that robust prompt engineering can be applied to generalist OS foundation models to significantly optimize performance in the absence of specialized fine-tuning. In this study, we present the OpenMedLM prompting platform, which applies robust prompt engineering techniques to the OS Yi 34B foundation model, absent fine-tuning, to achieve OS SOTA results on the four medical benchmarks we evaluated: MedQA, MedMCQA, PubMedQA, and the medical-subset of the MMLU. By employing a range of prompting techniques, including few-shot prompting (Brown et al., 2020), chain-of-thought (CoT) prompting (Wei, Wang, et al., 2022), and self-consistency (Wang et al., 2022), our OpenMedLM prompting platform when used with the OS Yi 34B foundation model outperformed Meditron on the MedQA (Fig. 1), MedMCQA, and the medical-subset of the MMLU benchmarks. We leveraged the prompting techniques similar to those presented by Nori et al. (Nori et al., 2023) with proprietary models to demonstrate that certain prompting strategies can achieve better accuracy on OS foundation models of known specifications, ensuring researchers, developers, and users are aware of the technical details behind the applications being created. We provide a thorough description of the prompting techniques utilized, as well as example prompts for each of these techniques. We further highlight the intermediate performance of each prompting strategy in an ablation study, and showcase the additive and synergetic effects of utilizing multiple prompting techniques on the OS SOTA performance of the OpenMedLM prompting platform.

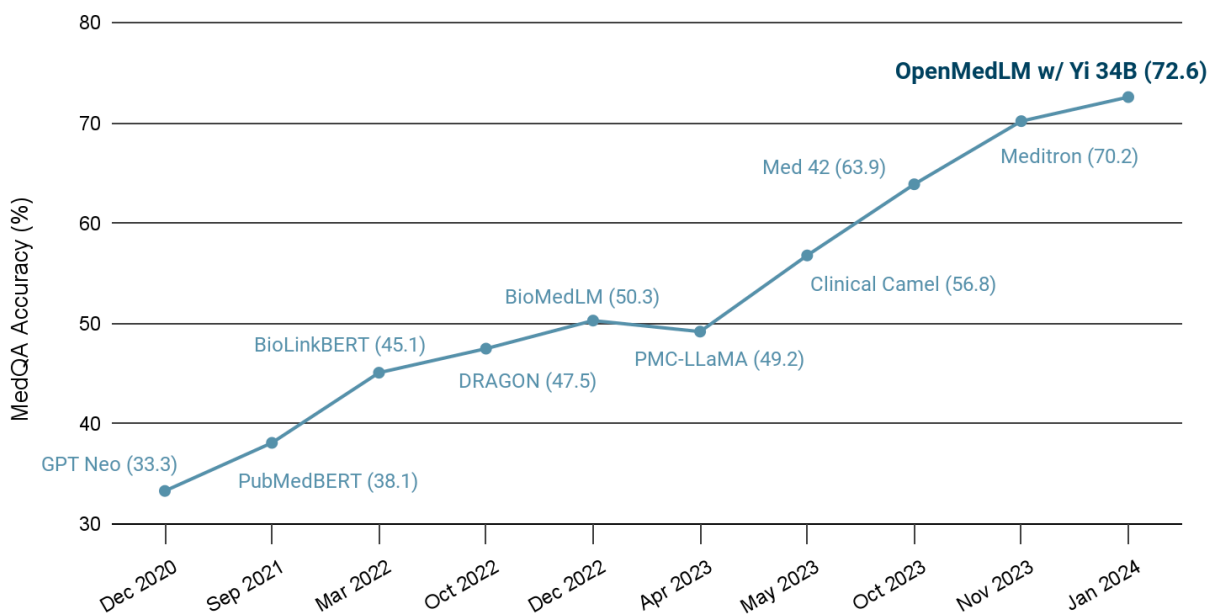

**Fig. 1: OpenMedLM Performance on the MedQA USMLE-style benchmark.** OpenMedLM achieves 72.6% accuracy on the MedQA dataset with the Yi 34B foundation model, surpassing all other OS models.

## 2. Methodology

### 2.1 Models

A range of OS foundation LLMs were tested to determine the model which achieved the best performance on the four medical benchmarks of interest (i.e., MedQA, MedMCQA, and the medical-subset of the MMLU benchmarks). We tested several LLMs ranging in size from 7B to 70B to evaluate baseline (zero-shot) performance across a varied range of models. Of these initial results, Yi 34B performed the best with zero-shot prompting, and thus was utilized for further experimentation with the OpenMedLM prompting platform. The OS Yi 34B foundation model is an LLM consisting of 34 billion parameters developed by 01.AI and was released in November 2023. The architecture of the model is derived from the architecture of the Llama model, but was trained independently of Llama and entails a set of model weights and parameters that are different from the Llama's weights and parameters ([Anon], 2023a).

### 2.2. (Evaluation) Datasets

For the LLM evaluation, we benchmark the performance on four multiple-choice, medical datasets. Table 1 provides an overview of the dataset details noted below:

MedQA (Jin et al., 2021) is another large-scale multiple-choice question answering dataset based on the USMLE, covering questions in English, simplified Chinese, and traditional Chinese. We present our evaluation benchmark on the English test subset of this dataset consisting of 1,273 questions.

MedMCQA (Pal et al., 2022) is a large-scale multiple-choice question answering dataset with a collection of more than 194,000 high-quality mock and historic exam questions from AIIMS and NEET-PG, two of the main Indian medical school entrance exams, covering as many as 2,400 topics and 21 subjects. We report our benchmarking on the "dev" subset of this dataset which contains 4,183 questions.

PubMedQA (Jin et al., 2019) is a dataset crafted to address research questions using abstracts from PubMed. The task in this dataset is to come up with "yes," "no," or "maybe" as the answer for each question in the dataset, given a context from PubMed abstracts. The questions could be evaluated either in a "reasoning-required" or a "reasoning-free" setting. In the reasoning-required setting, context from abstracts is provided along with the question and in the reasoning-free setting, a long-form answer containing the explanation of the abstracts is provided. We present our evaluation benchmark on 500 test questions from this dataset using the reasoning-required setting.

Medical-Subset of the MMLU (Hendrycks et al., 2020), where the entire MMLU dataset encompasses 57 topics from various fields like STEM, humanities, social sciences and other specialized areas designed to identify the strengths and weaknesses of language models in various domains. For our evaluation benchmarking, we use nine medically and clinically relevant subsets of topics including high school biology, college biology, college medicine, professional medicine, medical genetics, virology, clinical knowledge, nutrition, and anatomy with a total of 1,785 unique questions.

**Table 1: Description and item counts of evaluated medical benchmarks.** The size of the test set for each benchmark is indicated, along with the source and type of questions presented in each benchmark.

| Dataset | Test Item Count | Description |
| --- | --- | --- |
| MedQA (4 Options) | 1,273 | Multiple-choice question answers on medical knowledge form US medical licensing examinations |
| MedMCQA | 4,183 | Multiple-choice question answers on medical knowledge from Indian medical entrance exams |
| PubMedQA | 500 | Yes/No/Maybe question answering based on PubMed abstracts |
| MMLU Medical-Subset | 1,785 | Multiple-choice question answers from a subset of nine medical and clinical categories from the MMLU dataset |

2.3 Prompt Engineering Techniques

Prompting refers to the method of providing specific inputs to an LLM in order to elicit a desired response. This input is known as a prompt and can vary from a simple question (e.g., zero-shot) to a complex instruction with multiple examples (e.g., CoT). An effective prompt can guide the model to generate results that are relevant, accurate, and contextually appropriate for a specific task. Prompt engineering is the process of testing and evaluating a range of prompts to identify a template or prompt which elicits the most accurate response from the LLM (Lim & Schmälzle, 2023; Meskó, 2023). We employ the following prompt engineering strategies in our evaluation of OS foundation LLMs on the various medical benchmark datasets.

Zero-Shot prompting refers to a prompting style in which a task or a question is presented to the model without providing any context or examples (Kojima et al., 2022). We provide a brief instruction accompanied by the multiple choice question and the available options.

Few-Shot prompting refers to a type of in-context learning (ICL) strategy (Akyürek et al., 2022; Brown et al., 2020) where multiple examples are provided as context in addition to an instruction. Our few-shot prompting strategies include an instruction and five randomly chosen examples of multiple choice questions along with their correct answers. For every question in the test set, the instruction and the five examples are added to the prompt prior to the question.

Chain-of-Thought (CoT) prompting involves the creation of prompts which include intermediate steps of reasoning and deduction in the prompt text (Wei, Wang, et al., 2022). For example, a CoT prompt may include a simulated thought process of how a physician might deduce the answer to a question based on the key points presented in the body of the question. In our evaluation strategy, we employ the CoT strategy along with the few-shot strategy by providing a brief explanation along with the question and the correct answer for each example used in the prompt. We use two different approaches in selecting the examples for such prompts: (i) randomly selected examples from the training set, and (ii) k-nearest neighbors (kNN)(Qamar et al., 2008) strategy of selecting the 5 most similar questions for each test set question from the training set questions, a method adopted from the approach described by Nori et al. (Nori et al., 2023). As seen in Fig. 2a, when employing a CoT approach, the model outputs an explanation in addition to the answer to the question.

Ensemble/Self-Consistency: For the kNN strategy described above in the context of the CoT prompting, we also employ the ensemble or self-consistency strategy where we run the prompt multiple times through the model and implement a majority voting strategy to get the answer (Wang et al., 2022). This approach ensures the model selects the same answer in multiple runs and increases confidence in the model's output. In order to make this process even more reliable, for every run, we shuffle the options of the test question.

```
*** INSTRUCTION  ***

### Question: {question 1} {options 1}
### Explanation: {chain of thought 1}
### Answer: {correct answer 1}

### Question: {question 2} {options 2}
### Explanation: {chain of thought 2}
### Answer: {correct answer 2}

### Question: {question 3} {options 3}
### Explanation: {chain of thought 3}
### Answer: {correct answer 3}

### Question: {question 4} {options 4}
### Explanation: {chain of thought 5}
### Answer: {correct answer 4}

### Question: {question 5} {options 5}
### Explanation: {chain of thought 5}
### Answer: {correct answer 5}

### Question: {test question} {test options}
### Explanation:
```

**a.**

```
### Question: A 67-year-old man with[...] (A) Medication regimen
(B) Otitis externa (C) Otitis media (D) Presbycusis
### Explanation:
The patient described in the scenario has been receiving[...]
### Answer: Hence, the answer is D
Output generated from model
```

**b.**

**Fig. 2: The overall template for prompting each question for each benchmark, with a sample answer. a**, **b**, The 5 few-shot examples varied between the random selection and the kNN variants of prompting, and the explanations were generated by GPT-4 (**a**). An example output of the Yi 34B model to a few-shot CoT input prompt, where the question (abbreviated from length) is from the MedQA dataset (**b**).

2.4 Implementation

When experiments were performed to evaluate performance, the prompting techniques were run sequentially starting with zero-shot prompting, where each subsequent prompting approach is added on to previous techniques as an ablation study as described below:

i. Zero-shot prompting was run solely with an instruction followed by the question and multiple choice options.

ii. Few-shot prompting was run with an instruction followed by random selection examples consisting of questions and corresponding answers from the dataset. The question to be answered with corresponding options followed the randomly selected examples.

iii. The CoT prompting with the random few-shot examples was presented with an instruction, followed by an example question with possible answers, and then an explanation (CoT reasoning) along with the correct answer, subsequently followed by the question to be answered and its possible answers, as shown in Fig. 2a. For the CoT prompts, we generated the CoT explanations using GPT-4 via the OpenAI API. Every question selected to be used as examples in the CoT prompts were prompted to GPT-4 for generating a CoT explanation along with the correct answer. Fig. 3 shows the template for generating the CoT explanation which was provided to GPT-4.

iv. The CoT prompting with the kNN-based selection of few-shot examples followed a similar template for generating explanations as the CoT with random few-shot examples. However, as explanations needed to be generated for each of the 5 examples for each question, a caveat was added for incorrect answers during generation of the CoT explanation. If the answer was incorrect, the question was resubmitted up to 3 times for the correct answer to be generated. In cases where GPT-4 was not able to generate a correct answer from 3 trials, the question was replaced with the next most similar question from the training set based on kNN output (as detailed in the *Evaluation Methods* subsection below).

```
You are a medical expert. Answer the following multiple choice
question from the medical domain based on following instructions.
1. Output a brief explanation summarizing and providing context to
the question under the heading 'Explanation' in about 5 sentences.
2. Select the correct option and provide the correct option under
the heading 'Answer'.
3. Always select one of the four options provided as the answer.
4. If the options are ambiguous or the question does not have enough
context, select the one that best answers the question.

### Question: {question}
### Options: {options}
```

**Fig. 3: The prompt fed to GPT-4 to generate the chain-of-thought (CoT) explanations for each few-shot example for both the random few-shot examples and the kNN-based few-shot examples.** For kNN CoT, the prompt was run up to 3 times per example to achieve a correct answer before the next most similar example was fed into GPT-4.

2.5 Evaluation Methods

Due to our focus on multiple choice Q&A benchmarks, we use accuracy as the evaluation metric to compare performance. During the inference for evaluation, we allowed a maximum of 5 tries for the model to generate a valid output. A valid output was defined as an output which contained an item that could be interpreted as an answer to a multiple choice question (e.g., "(A)"). If the model did not generate a valid output after 5 tries, the answer was counted as incorrect and would reflect as such in the final computation of accuracy for the benchmark. One key factor in generating valid outputs, specifically for generating answers to CoT prompting, is the maximum number of tokens output by the model. Fig. 2b shows an example of an output which employs a CoT approach. The explanation is followed by the multiple choice answer which is clearly delineated with the appropriate heading. As the maximum number of tokens for the model also affects the GPU memory usage and the time for evaluation, we selected a value for the maximum number of tokens empirically to mimic the average length of the tokens in the CoT explanations of the questions in the examples used in the prompt. The instructions in the prompts were also selected empirically to optimize performance, as follows. A series of prompt instructions were tested in order to determine a set of instructions which maximized performance. We tested instructions including customized medical instructions, standard Q&A instructions, and instructions presented elsewhere in the literature such as the prompt examples provided by Singhal et. al with the Med-PaLM specialized model (Singhal, Azizi, et al., 2023).

As a variation of the CoT examples, we implemented a kNN method, based on the approach described by Nori et al. (Nori et al., 2023), to select the most similar questions from the respective training sets for every test question. The kNN approach employed by Medprompt runs each test prompt through a kNN algorithm to find the 5 nearest neighbors to the test question in the dataset's training split, creating unique prompts for each test question to best encompass the context of that particular test question. Additionally, the ensemble voting scheme was implemented, which runs each prompt through the model 5 times and selects the answer most commonly output during the 5 runs. Similar to the Medprompt approach, for each of the 5 inference runs, the multiple choice options were randomly shuffled to create variation in which option corresponded to the correct answer. Each of the 5 inference runs was performed with a decoding temperature of 0.4. The decoding temperature is a model parameter which influences how deterministic the output is. Lower temperature values indicate a more deterministic output and higher temperatures correlate with more diverse or creative outputs. For each benchmark, an ablation experiment was performed to evaluate the model's performance on each prompting technique individually prior to adding on additional prompting techniques. The ablation experiment was performed on all benchmarks (Table 2).

Following the ablation study, results were computed for each prompting strategy, as well as for the complete OpenMedLM prompting platform in order to determine the accuracy of the robust prompt engineering approach as applied to the OS Yi 34B foundation model.

3. Results

Through the combination of all sub-components of the OpenMedLM prompting platform, we showcase OS SOTA performance on 3 of the 4 medical Q&A benchmarks evaluated in this study. Table 2 highlights the performance of the Yi 34B foundation model with the complete OpenMedLM prompting platform (i.e., employing all prompting strategy components of OpenMedLM) compared against the performance of the Meditron 70B specialized model, which achieved the previous SOTA for OS LLMs. Table 2 further shows the breakdown of results of each prompting strategy in an ablation experiment.

Table 2: Performance of top-performing OS models on multiple-choice medical benchmarks. OpenMedLM delivers state-of-the-art results on 3 of the 4 primary benchmarks with the Yi 34B foundation model. Results for the specialized Meditron model were sourced from Chen et. al. (Chen et al., 2023) (ZS = zero-shot; FS = few-shot; CoT = chain-of-thought; kNN = k-nearest neighbors).

|  | Benchmark Accuracy | | | |
| --- | --- | --- | --- | --- |
| **Model** | **MedQA (4 Options)** | **MedMCQA** | **PubMedQA** | **MMLU - Medical** |
| | *ZS* | | | |
| Yi 34B | 58.4 | 55.9 | 53.4 | 72.6 |
| Meditron 70B | 65.4 | 65.1 | 80.0 | 73.6 |
| | *Random FS* | | | |
| Yi 34B | 61.5 | 58.0 | 57.0 | 77.3 |
| | *CoT* | | | |
| Yi 34B (FS) | 66.6 | 59.2 | 68.2 | 75.2 |
| Meditron 70B (ZS) | 67.8 | 63.2 | 81.0 | 74.9 |
| *kNN FS CoT* | | | | |
| Yi 34B | 69.5 | 65.7 | 72.4 | 78.7 |
| *Ensemble/Self-Consistency* | | | | |
| Yi 34B (kNN FS w/ Shuffling) | **72.6** | **68.3** | 77.3 | **81.6** |
| Meditron 70B (Random FS) | 70.2 | 66.0 | **81.6** | 77.6 |

The ablation experiment highlights the contribution of each sub-component of the OpenMedLM prompting platform to the overall results of the model. Fig. 2 showcases the effect of each sub-component of OpenMedLM for the MedQA benchmark, by comparison with the compounding effect of each prompting technique employed by Meditron to achieve their model's top accuracy. While Meditron outperforms the Yi 34B model with zero-shot prompting (i.e., 65.4% vs. 58.4% accuracy, respectively), the combination of all components of the OpenMedLM prompting platform results in a prompt engineering approach that enables the Yi 34B foundation model to outperform the specialized Meditron model.

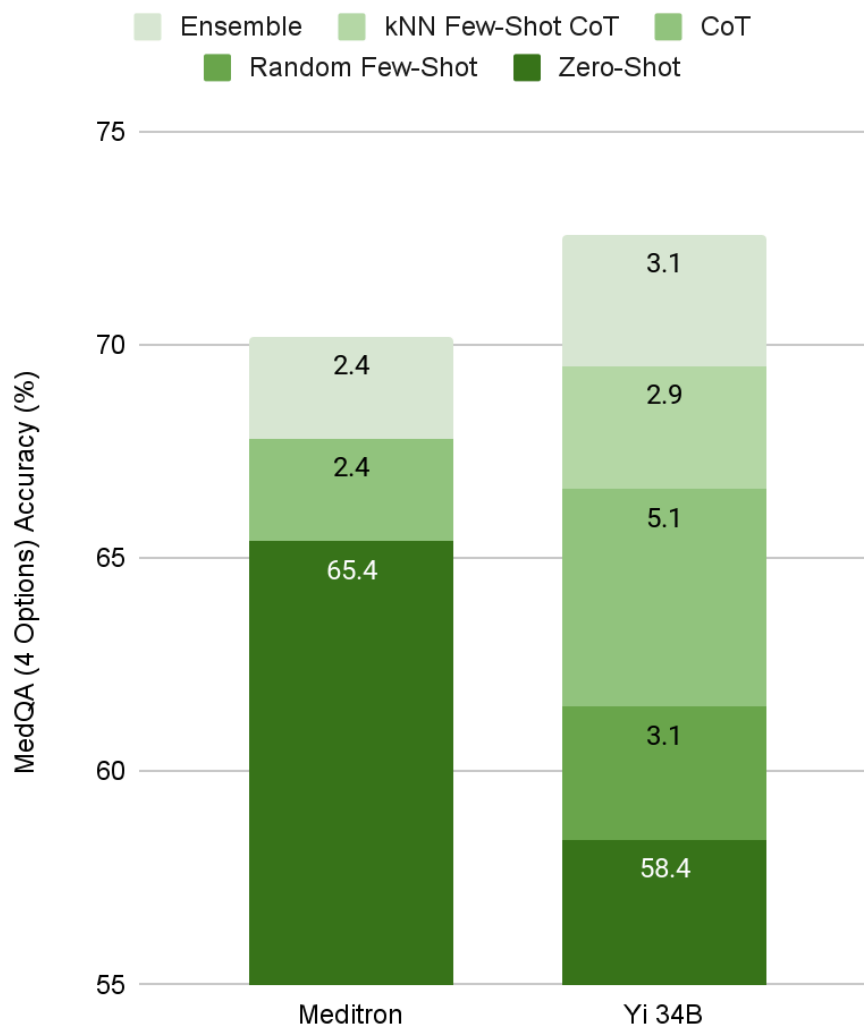

**Fig. 4: Ablation study showing the contribution of different components of the OpenMedLM prompting platform.** While the accuracy of zero-shot prompting a fine-tuned specialized model (Meditron 70B) on the MedQA dataset is higher than the accuracy of the Yi 34B foundation model with the same zero-shot prompting, through the addition of all components of OpenMedLM, Yi 34B achieves a higher accuracy than Meditron on the MedQA dataset.

*Zero-Shot*: The baseline performance of the Yi 34B foundation model absent any prompt engineering techniques achieves an accuracy of 58.4% on the MedQA dataset. This was lower than the 65.4% achieved by the specialized Meditron model with a zero-shot prompting approach.

*Random Few-Shot*: Following the implementation of random few-shot prompting, the performance of the Yi 34B foundation model increased by 3.1% to deliver an accuracy of 61.5%. The few-shot prompts selected for this experiment were a random selection of prompts from the dataset and were constant for all questions in the MedQA benchmark.

*Random Few-Shot with CoT*: CoT explanations were utilized for each of the few-shot examples and added as the input prompt to the Yi 34B foundation model. This prompting strategy resulted in the largest surge in performance, contributing to a 5.1% increase that delivered an accuracy of 66.6% for the Yi 34B foundation model. In contrast, a CoT approach improves the performance of the Meditron specialized model by only 2.4%, although it should be noted that Meditron solely utilizes zero-shot CoT.

*kNN Few-Shot with CoT*: To further test the importance of the specific examples utilized in few-shot prompting, a kNN algorithm was implemented for each question in each benchmark for the evaluation of the Yi 34B foundation model. The kNN algorithm identified the 5 most similar training questions to each test question and generated a CoT explanation for each of those examples to construct the prompt. The kNN few-shot with CoT approach led to a 2.9% improvement in accuracy on MedQA to deliver an accuracy of 69.5% for the Yi 34B foundation model.

*Ensemble/Self-Consistency*: The final prompting strategy that we employed utilized a self-consistency approach to develop a majority voting scheme for answer selection. The voting scheme provided an additional 3.1% improvement in performance which led to the achievement of an overall total accuracy of 72.6% for the Yi 34B foundation model.

While the implementation of CoT explanations (kNN Few-Shot with CoT) provided the largest enhancement of model performance, each sub-component of the OpenMedLM prompting platform delivers at least a 2.9% improvement in accuracy, thus contributing to the total performance improvement of 14.2% over zero-shot prompting on the Yi 34B foundation model. When evaluating the ablation study on the MedMCQA and PubMedQA benchmarks, we see similar results with each sub-component of the OpenMedLM prompting platform providing additional improvement in accuracy. As with the MedQA benchmark, the largest performance increase on the PubMedQA benchmark was seen with the Random Few-Shot CoT approach, contributing 11.2% increase in performance. The overall performance of the Yi 34B foundation model with the OpenMedLM prompting platform was 77.3% on the PubMedQA benchmark, falling short of the 81.6% achieved by Meditron. This was the only benchmark on which the Yi 34B foundation model performed worse than Meditron. However, for the MedMCQA benchmark, the largest performance boost came from addition of the kNN Few-Shot CoT, leading to a 6.5% increase in performance compared to the Random Few-Shot CoT. The overall accuracy of the Yi 34B foundation model with OpenMedLM on MedMCQA was 68.3%, a 2.3% increase over the performance of Meditron.

In contrast to the other three datasets, the sequential addition of the OpenMedLM prompting strategies for the medical-subset of the MMLU dataset displayed a synergistic effect rather than an additive one. While the combination of all sub-components of OpenMedLM resulted in a substantial improvement over the zero-shot baseline, the addition of CoT to the Random Few-Shot approach decreased performance by 2.1%. The further addition of kNN Few-Shot CoT then boosted performance by 3.5%, exceeding the performance from Random Few-Shot alone. For the medical-subset of the MMLU, the largest increase in performance came from the initial Random Few-Shot prompting strategy, resulting in a 4.7% increase in performance. The overall accuracy of the Yi 34B foundation model with OpenMedLM was 81.6%, surpassing the 77.6% achieved by Meditron. Across all four benchmarks, the addition of the OpenMedLM prompting platform improved accuracy by at least 9% compared to the zero-shot baseline.

## 4. Discussion

We present OpenMedLM, a prompting platform which enables SOTA results for OS foundation LLMs on common medical benchmarks. OpenMedLM facilitates OS SOTA level performance solely by utilizing robust prompt engineering on generalist OS foundation LLMs. Through a series of additive and synergistic prompting techniques, OpenMedLM enables performance superior to previous OS SOTA on three of the evaluated medical benchmarks for the Yi 34B foundation model we employed, which showcases the potential of generalist OS foundation models to perform highly specialized tasks without the costly challenges of requiring a highly specialized training or fine-tuning dataset that are necessary to develop specialized models.

We leveraged the combination of several proven prompt engineering techniques into OpenMedLM in order to achieve OS SOTA results, including the first reported instance of a generalist OS foundation LLM achieving over 80% accuracy on the medical subset of the MMLU benchmark. The accuracy results displayed in Table 2 demonstrate that the prompting techniques we employed offer a combination of additive and synergistic benefits for different benchmarks. On the MedQA, MedMCQA, and PubMedQA benchmarks, our ablation study shows that the addition of each prompting technique progressively improves performance of the model, with the combination of (1) CoT with kNN-based few-shot prompting and (2) shuffled-option self-consistency resulting in the highest performance overall. In particular, this prompting technique of combining very specific types of few-shot prompting and self-consistency achieved the highest accuracy on the MedQA and MedMCQA benchmarks amongst generalist OS foundation LLMs to-date. This is highlighted in Fig. 4, which shows that the addition of each prompting technique offered at least a 2.9% boost in accuracy on MedQA, and the combination of all four prompting techniques delivers SOTA performance for the generalist OS model. The results of the ablation experiment on the medical-subset of the MMLU benchmark show a combination of additive and synergistic effects. CoT prompting in combination with a kNN-based selection of the few-shot examples offers an increase in performance compared to a few-shot baseline, while CoT with a random selection of few-shot examples reduces performance. As with the other benchmarks, the combination of these prompting strategies with shuffled-option self-consistency delivers the best performance for the generalist OS model at 81.6% accuracy. The order in which the prompting techniques are employed is consequential to the performance, indicating that the prompting strategies in OpenMedLM are not statistically independent. The sequence to add each successive prompting technique was motivated by the preceding techniques that were employed. For example, few-shot prompting with CoT extends from regular few-shot prompting by adding the CoT explanations to each of the few-shot examples. Therefore, with the final prompt, all techniques described were utilized in the model input.

Results such as those presented by Nori et al. (Medprompt) (Nori et al., 2023) have showcased the ability to achieve emergent properties through prompting alone in very large proprietary foundation models lacking publicly available specifications, such as GPT-4. While it is now established that emergent properties are present in these very large models, it has not been shown to date whether these types of properties can emerge at the top end of OS models which have intrinsically fewer parameters than many of the proprietary LLMs. In this paper, we showcase that some medical Q&A-related emergent properties are also apparent in OS foundation models and that prompting alone can outperform intensive fine-tuning when performing specialized healthcare tasks.

As the research into the use of LLMs for healthcare applications continues, it is important to understand the capabilities of both generalist (foundation) and specialized (fine-tuned) models in order to determine the approach that can lead to the most accurate, complete, and unbiased generated responses for physicians and patients. While extended pre-training and fine-tuning has been shown to provide substantial improvements in performance compared to generalist foundation models alone, the computational costs, complexity of data, and potential for catastrophic forgetting adds challenges that may render the development of SOTA-performing specialized models inaccessible to all but those with very large budgets. Achieving best performing results on specialized tasks by employing generalist OS foundation models (as opposed to generalist or specialized proprietary models) highlights the need to continue research into better understanding the full extent of the capabilities of these OS foundation models and leverage the benefits they provide with regards to accessibility and generalizability. Research into LLMs has shown that large models have potential emergent abilities which may not have been thought of when the model was trained (Wei, Tay, et al., 2022). The ability to achieve high accuracy on medical Q&A with foundation models highlights the emergence of properties for healthcare specific tasks (e.g., medical Q&A-related emergent properties) which were previously assumed to require fine-tuning. The ability of OpenMedLM to bring out emergent properties in OS models substantiates the need for further research into the full capabilities of generalist OS foundation LLMs on medical tasks. In addition, further studies should be performed to analyze the combination of fine-tuning in combination with advanced prompting techniques to determine if the findings in this study also extend to specialized models derived from OS foundation models.

4.1 Limitations

While the studies presented in this paper achieve SOTA results for OS foundation LLMs, there is still a significant amount of research and experimentation required prior to these models being applicable in real-world scenarios. The benchmarks evaluated in this study are limited to Q&A from an academic perspective and may not encompass the types of questions a physician or patient may ask in a true clinical setting. Additionally, all of the questions evaluated by each benchmark are multiple choice questions requiring the model to choose a single answer out of multiple options. While there are real-world healthcare encounters which involve the selection of a single answer out of a few options, most scenarios would still require open-ended responses that go beyond selecting a response based on multiple choices. Further, the complexity of real-world scenarios may increase the chance of hallucinations, incomplete, or biased responses, as the LLM would have to output a more sophisticated response that may include reasoning, evidence, or next-steps, which are all beyond a simple answer to a question. We think that further experimentation will demonstrate that OS LLMs have the potential to answer more complex questions both at a physician-centric and patient-centric level and will be utilized in real-world scenarios in the near future.

## 5. Conclusions

We present OpenMedLM as a way to showcase the ability of generalist OS foundation LLMs to achieve strong performance on medical tasks, which can offer benefits for the use of LLMs in healthcare applications that are not met by proprietary models, such as accessibility and customizability. The prompting strategies as well as the models are publically available, and we hope that our results can encourage further research on how to use prompt engineering in new and creative ways to enhance the ability of the LLMs to perform medical tasks. We think that a combination of strategies can be employed to further improve model performance with the end goal of developing practical and accessible tools that can be utilized in real-world clinical scenarios to provide additional resources for physicians, therefore enabling better health outcomes for patients. Furthermore, we think that integrating LLM capabilities with other medical AI algorithms, which already show promise in diagnostics and treatment delivery for example (Adelson et al., 2023; Maharjan et al., 2023), can provide for extremely powerful yet accessible tools that present the additional advantage of ease of implementation in healthcare workflows.

**Author contributions:** JM and AG contributed to the experimental design, execution, and writing of the Methods section; AG was lead writer for the manuscript; NPS, RT, QQ, and RD contributed to conceptual design of the experiments; LC, MS, and QQ contributed to the overall execution of experiments; MC and QQ contributed to the writing, editing, and overseeing the development of the manuscript; GB contributed to editing of the manuscript. All authors read and approved the final manuscript.

**Acknowledgements:** There are no external funding sources to acknowledge.

**Competing interests:** All authors are employees or contractors of Montera, Inc. dba Forta.

**Data availability:** The benchmarks utilized in the study are publically available and referenced in the Methods section of the manuscript. The Yi 34B model is open source and available under the Apache 2.0 license for academic use. The model is cited throughout the manuscript and can be accessed through the link below.

**Code availability:** The code to run the Yi 34B model is available at https://github.com/01-ai/Yi. The other code required to produce the findings of the study is available from the corresponding author under reasonable request.

**Correspondence and requests for materials should be addressed to Qingqing Mao.**